\documentclass[11pt,a4paper]{article}
\usepackage[hyperref]{emnlp-ijcnlp-2019}
\usepackage{times}
\usepackage{latexsym}
\usepackage{url}

\usepackage{helvet}  
\usepackage{courier}  
\usepackage{graphicx}  
\usepackage{CJK}
\usepackage{color}
\usepackage{multirow}
\usepackage{multicol}
\usepackage{enumitem}
\usepackage{amsmath}
\usepackage{amssymb}
\usepackage{textcomp}
\usepackage{subfig}
\usepackage{array}
\usepackage{xspace}
\usepackage{wasysym}
\usepackage{arydshln}
\usepackage{balance}

\aclfinalcopy 

\definecolor{forestgreen}{rgb}{0.13, 0.55, 0.13}
\definecolor{ao}{rgb}{0.0, 0.5, 0.0}
\definecolor{napiergreen}{rgb}{0.16, 0.5, 0.0}
\definecolor{brown}{rgb}{0.59, 0.29, 0.0}

\newcolumntype{L}[1]{>{\raggedright\arraybackslash}p{#1}}
\newcolumntype{C}[1]{>{\centering\arraybackslash}p{#1}}
\newcolumntype{R}[1]{>{\raggedleft\arraybackslash}p{#1}}

\newcommand{\ie}{\emph{i.e.,}\xspace}

\DeclareMathOperator*{\argmax}{arg\,max}

\title{One Model to Learn Both: Zero Pronoun Prediction and Translation}

\author{
Longyue Wang \\ Tencent AI Lab \\ {\tt vinnylywang@tencent.com} \And
Zhaopeng Tu \\ Tencent AI Lab \\ {\tt zptu@tencent.com} \AND
Xing Wang \\ Tencent AI Lab \\ {\tt brightxwang@tencent.com} \And
Shuming Shi \\ Tencent AI Lab \\ {\tt shumingshi@tencent.com}
}

\date{}

\begin{document}
\maketitle
\begin{abstract}
Zero pronouns (ZPs) are frequently omitted in pro-drop languages, but should be recalled in non-pro-drop languages.
This discourse phenomenon poses a significant challenge for machine translation (MT) when translating texts from pro-drop to non-pro-drop languages. In this paper, we propose a unified and discourse-aware ZP translation approach for neural MT models. Specifically, we jointly learn to predict and translate ZPs in an end-to-end manner, allowing both components to interact with each other. 
In addition, we employ hierarchical neural networks to exploit discourse-level context, which is beneficial for ZP prediction and thus translation. Experimental results on both Chinese$\Rightarrow$English and Japanese$\Rightarrow$English data show that our approach significantly and accumulatively improves both translation performance and ZP prediction accuracy over not only baseline but also previous works using external ZP prediction models. Extensive analyses confirm that the performance improvement comes from the alleviation of different kinds of errors especially caused by subjective ZPs.

\end{abstract}

\section{Introduction}

Zero anaphora is a discourse phenomenon, where pronouns can be omitted when they are pragmatically or grammatically inferable from intra- and inter-sentential context~\cite{li1979third}. However, translating such implicit information (i.e. zero pronoun, ZP) poses various difficulties for machine translation (MT) in terms of completeness and correctness. Although neural models are getting better at learning representations, it is still difficult to implicitly learn complex ZPs in a general model. Actually, ZP prediction and translation need to not only understand the semantics or intentions of a single sentence, but also utilize its discourse-level context.


Two technological advances in the field of ZP and MT, have seen vast progress over the last decades, but they have been developed very much in isolation. Early studies~\cite{chung2010effects,Nagard:2010:ACL,xiang2013enlisting} fed MT systems with the results of ZP prediction models, which are trained on a small-scale and non-homologous data compared to MT models.
To narrow the data-level gap,~\newcite{Wang:2016:NAACL} proposed an automatic method to annotate ZPs by utilizing the parallel corpus of MT. The homologous data for both ZP prediction and translation leads to significant improvements on translation performances for both statistical MT~\cite{Wang:2016:NAACL} and neural MT models~\cite{Wang:2018:AAAI}.
However, such approaches still require external ZP prediction models, which have a low accuracy of 66\%. The numerous errors of ZP prediction errors will be propagated to translation models, which leads to new translation problems.
In addition, relying on external ZP prediction models in decoding makes these approaches unwieldy in practice, due to introducing more computation cost and pipeline complexity.


In this work, we try to further bridge the model-level gap by jointly modeling ZP prediction and translation.
Joint learning has proven highly effective on alleviating the error propagation problem, such as joint parsing and translation~\cite{Liu:2010:COLING}, as well as joint tokenization and translation~\cite{Xiao:2010:COLING}.
Similarly, we expect that ZP prediction and translation could interact with each other: prediction offers more ZP information beyond 1-best result to translation and translation helps prediction resolve ambiguity.
Specifically, we first cast ZP prediction as a sequence labeling task with a neural model, which is trained jointly with a standard neural machine translation (NMT) model in an end-to-end manner. We leverage the auto-annotated ZPs to supervise the learning of ZP prediction component, which releases the reliance on external ZP knowledge in decoding phase.

In addition, previous studies revealed that discourse-level information can better tackle ZP resolution, because around 23\% of ZPs appear two or more sentences away from their antecedents~\cite{zhao2007identification,chen2013chinese}.
Inspired by these findings, we exploit inter-sentential context to further improve ZP prediction and thus translation. 
Concretely, we employ hierarchical neural networks~\cite{Sordoni2015A,Wang:2017:EMNLP} to summarize the context of previous sentences in a text, which is integrated to the joint model for ZP prediction.

We validate the proposed approach on the widely-used data for ZP translation~\cite{Wang:2018:AAAI}, which consist of 2.15M Chinese--English sentence pairs.
Experimental results show that the joint model indeed improves performances on both ZP prediction and translation. Incorporating discourse-level context further improves performances, and outperforms ther external ZP prediction model~\cite{Wang:2018:AAAI} by +2.29 BLEU points in translation and +11\% in prediction accuracy.
Experimental results on a further Japanese--English translation task show that our model consistently outperforms both the baseline and the external ZP prediction model, demonstrating the universality of the proposed approach.

The key contributions of this paper are:
\begin{enumerate}
  \item We propose a single model to jointly learn ZP prediction and translation, which improves performances on both tasks by allowing the two components to interact with each other.
  \item Our study demonstrates the effectiveness of discourse-level context for ZP prediction.
  \item Based on our manually-annotated testset, we conduct extensive analyses to assess ZP prediction and translation.
 \end{enumerate}

\section{Background}
\label{sec:2}

\subsection{Zero Pronoun}
\label{sec:2.1}

\begin{CJK}{UTF8}{gbsn}
\begin{table}[t]
\renewcommand\arraystretch{1.1}
\centering
    \begin{tabular}{R{0.55cm}|L{6.3cm}}
        \hline
    	Inp. & 等 我 搬进来，{\bf \color{red}(我)} 能 买 台 电视 吗？\\
    	Ref. & Can {\bf \color{red}I} get a TV when I move in? \\
    	Out. & 
    	When I move in {\bf \color{blue}to buy} a TV.\\ \hline
    	Inp. & 这块 \underline{\color{brown}蛋糕} 很 美味！你 烤 的 {\bf \color{red}(它)} 吗？\\
    	Ref. & The cake is very tasty! Did you bake {\bf \color{red} it}? \\
    	Out. & The cake is delicious!  {\bf \color{blue}Are you baked}? \\ \hline
    \end{tabular}
	\caption{Examples of ZPs and translations where words in brackets are ZPs that are invisible in decoding and underlined words are antecedents of anaphoric ZPs. This leads to problems for NMT in respect of completeness (first case) and correctness (second case). ``Inp.'' and ``Ref.'' indicate Chinese input and English translation, respectively. ``Out.'' represents the output of a NMT model.}
	\label{tab-zpexample}
\end{table}
\end{CJK}

In pro-drop languages such as Chinese and Japanese, ZPs occur much more frequently compared to non-pro-drop languages such as English~\cite{zhao2007identification}.
\begin{CJK}{UTF8}{gbsn}
As seen in Table~\ref{tab-zpexample}, the subject pronoun (``我'') and the object pronoun (``它'') are omitted in Chinese sentences (``Inp.'') while these pronouns are all compulsory in their English translations (``Ref.''). This is not a problem for human beings since we can easily recall these missing pronoun from the context. Taking the second sentence for example, the pronoun ``它'' is an anaphoric ZP that refers to the antecedent (``蛋糕'') in previous sentence, while the non-anaphoric pronoun ``我'' 
can still be inferred from the whole sentence. The first example also indicates the necessity of intra-sentential information for ZP prediction. \end{CJK} 

However, ZP poses a significant challenge for translation models from pro-drop to non-pro-drop languages, where ZPs are normally omitted in the source side but should be generated overly in the target side. As shown in Table~\ref{tab-zpexample}, even a strong NMT model fails to recall the implicit information, which lead to problems like {\em incompleteness} and {\em incorrectness}. The first case is translated into ``When I move in to buy a TV'', which makes the output miss subject element ({incompleteness}). The second case is translated into ``Are you baked?'', while the correct translation should be ``Did you bake it?'' ({incorrectness}). 


\subsection{Bridging Data Gap Between ZP Prediction and Translation}

Recent efforts have explored ways to bridge the gap of ZP prediction and translation~\cite{Wang:2016:NAACL,Wang:2018:AAAI,Wang:2018:EMNLP} by training both models on the homologous data. The pipeline involves two phases, as described below.

\paragraph{Translation-Oriented ZP Prediction}
\begin{CJK}{UTF8}{gbsn}
Its goal is to recall the ZPs in the source sentence (i.e. pro-drop language) with the information of the target sentence (i.e. non-pro-drop language) in a parallel corpus. Taking the second case (assuming that Inp. and Ref. are sentence pair in a parallel corpus) in Table~\ref{tab-zpexample} for instance, the ZP ``它 (\textit{it})'' is dropped in the Chinese side while its equivalent ``it'' exists in the English side. It is possible to identify the ZP position (between ``的'' and ``吗'') by alignment information, and then recover the ZP word ``它'' by a language model (scoring all possible pronoun candidates and select the one with the lowest perplexity).
~\newcite{Wang:2016:NAACL} proposed a novel approach to automatically annotate ZPs using alignment information from bilingual data, and the auto-annotation accuracy can achieve above 90\%.
Thus, a large amount of ZP-annotated sentences were available to train an external ZP prediction model, which was further used to annotate source sentences in test sets during the decoding phase. They integrated the ZP predictor into SMT and showed promising results on both Chinese--English and Japanese--English data.

However, their neural-based ZP prediction model still produce low accuracies on predicting ZPs, which is 66\% in F1 score. 
This is a key problem for the pipeline framework, since numerous errors would be propagated to the subsequent translation process. 
\end{CJK}

\paragraph{Translation with ZP-Annotated Data}
An intuitive way to exploit the annotated data is to train a standard NMT model on the annotated parallel corpus, which decodes the input sentence annotated by the external ZP prediction model.~\newcite{Wang:2018:AAAI} leveraged the encoder-decoder-reconstructor framework~\cite{Tu:2017:AAAI} for this task, which reconstructs the intermediate representations of NMT model back to the ZP-annotated input. The auxiliary loss on ZP reconstruction can guide the intermediate representations to learn critical information relevant to ZPs. However, their best model still needs external ZP prediction at decoding time. In response to this problem, \newcite{Wang:2018:EMNLP} leveraged the prediction results of the ZP positions, which have relatively higher accuracy (e.g. 88\%). Accordingly, they jointly learn the partial ZP prediction (\ie predict the ZP word given the externally annotated ZP position) and ZP translation.

In this work, we follow this direction with the encoder-decoder-reconstructor framework, and show our approach outperforms both strategies of using externally annotated data. 


\section{Approach}
\label{sec:approach}


In this study, we propose a joint model to learn ZP prediction and translation, which can be further improved by leveraging discourse-level context.
\begin{itemize}
    \item {\em Joint ZP Prediction and Translation} (Section~\ref{sec:3.1}) We cast ZP prediction as a sequence labeling problem, which can be trained together with ZP translation model in an end-to-end manner. This releases the reliance on external ZP prediction models (e.g. 66\% or 88\% accuracy), since no ZP-annotated sentence is required any more in decoding. Instead, only the high-quality annotated bilingual data (e.g. 93\% accuracy) are needed.
    \item {\em Discourse-Aware ZP Prediction} (Section~\ref{sec:3.2}) We further improve ZP prediction and thus its translation with discourse-level context, which is summarized by hierarchical neural networks. The contextual representation is integrated into the reconstructor, based on which ZP prediction is conducted.
\end{itemize}

\subsection{Joint ZP Prediction and Translation}
\label{sec:3.1}

\begin{figure}[t]
\centering
\includegraphics[width=0.48\textwidth]{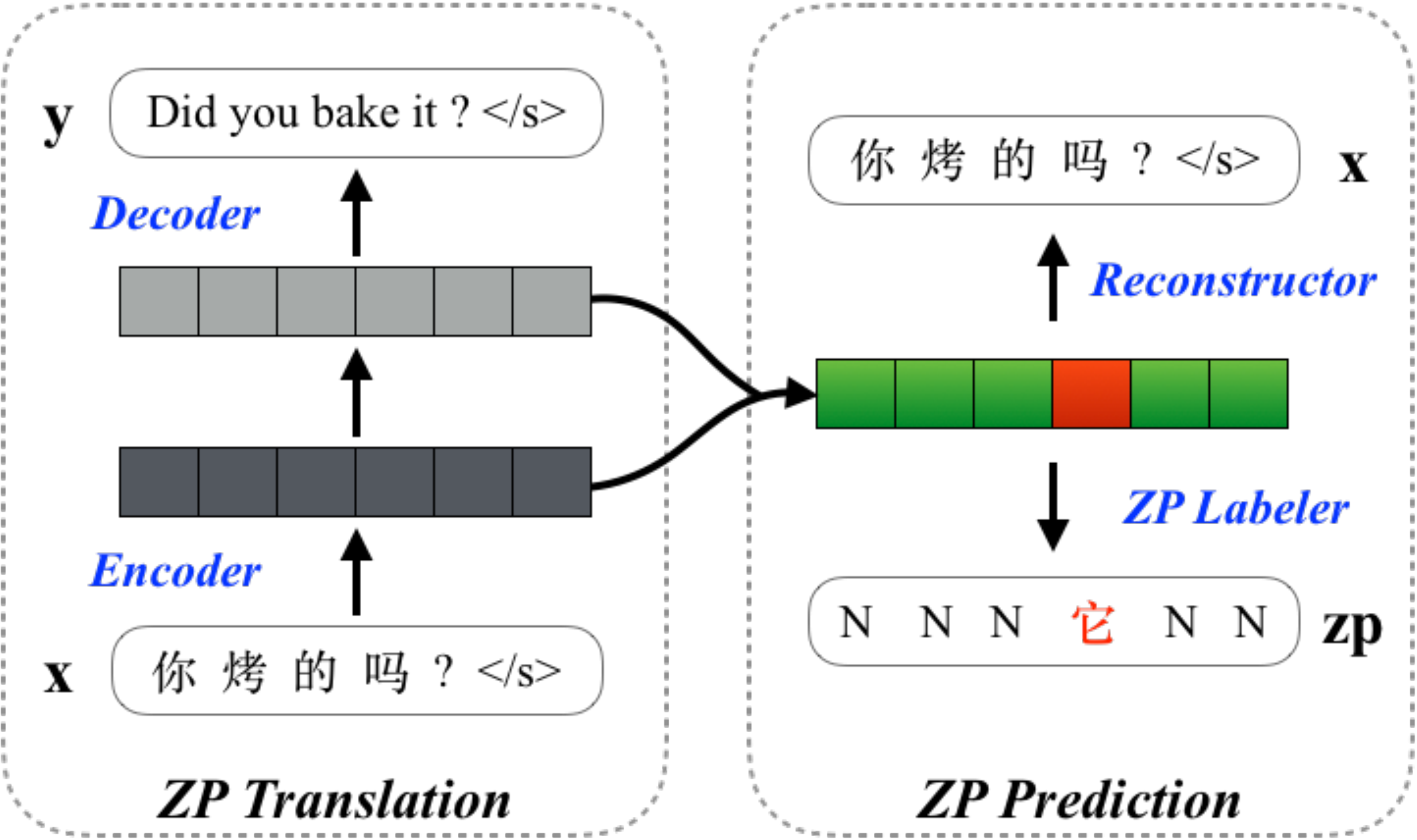}
\caption{Architecture of the joint ZP prediction and translation model, in which ZP prediction is casted as a sequence labelling problem.}
\label{fig-architecture}
\end{figure}

Figure~\ref{fig-architecture} illustrates the architecture of the joint model, which consists of two main components. The ZP translation component is a standard encoder-decoder NMT model, while an additional reconstructor is introduced for ZP prediction. 
To guarantee the reconstructor states contain enough information for ZP prediction, the reconstructor reads both the encoder and decoder states and the reconstruction score is computed by
\begin{eqnarray}
R({\bf \hat{x}}|{\bf h}^{enc}, {\bf h}^{dec})  = \prod_{t=1}^{T} g_r({\hat{x}}_{t-1}, {\bf h}^{rec}_t, \hat{\bf c}^{enc}_t, \hat{\bf c}^{dec}_t) \nonumber
\label{eqn:rec}
\end{eqnarray}
where ${\bf h}^{rec}_t$ is the hidden state in the reconstructor:
\begin{eqnarray}
{\bf h}^{rec}_t &=& f_r(\hat{x}_{t-1}, {\bf h}^{rec}_{t-1}, \hat{\bf c}^{enc}_t, \hat{\bf c}^{dec}_t)
\end{eqnarray}
Here $g_r(\cdot)$ and $f_r(\cdot)$ are respective softmax and activation functions for the reconstructor.
The context vectors $\hat{\bf c}^{enc}_t$ and $\hat{\bf c}^{dec}_t$ are the weighted sum of ${\bf h}^{enc}$ and ${\bf h}^{dec}$, and the weights are calculated by two interactive attention models:
\begin{eqnarray}
    \hat{\alpha}^{enc} &=& \textsc{Att}_{enc}(x_{t-1}, {\bf h}^{rec}_{t-1}, {\bf h}^{enc}) \\
    \hat{\alpha}^{dec} &=& \textsc{Att}_{dec}(x_{t-1}, {\bf h}^{rec}_{t-1}, {\bf h}^{dec}, \hat{\bf c}^{enc}_t)
\end{eqnarray}
The interaction between two attention models leads to a better exploitation of the encoder and decoder representations \cite{Wang:2018:EMNLP}.

\begin{CJK}{UTF8}{gbsn}
\paragraph{ZP Prediction as Sequence Labelling}
We cast ZP prediction as a sequence labelling task, where each word is labelled if there is a pronoun missing before it. Given the input ${\bf x}=\{{x}_1, {x}_2, \dots, {x}_T\}$ with the last word $x_T$ being the end-of-sentence tag ``
$\langle$eos$\rangle$'',\footnote{We introduce ``
$\langle$eos$\rangle$'' to cover the case that a pronoun is missing at the end of a sentence.} the output to be labelled is a sequence of labels ${\bf zp} = \{{zp}_1, {zp}_2, \dots, {zp}_T\}$ with ${zp}_t \in \{N\} \cup \mathbb{V}_{zp}$. Among the label set, ``$N$'' denotes no ZP, and $\mathbb{V}_{zp}$ is the vocabulary of pronouns.\footnote{We employ the pronoun vocabulary used in~\newcite{Wang:2016:NAACL}, which contains 30 distinct Chinese pronouns.} Taking Figure~\ref{fig-architecture} as an example, the label sequence ``N N N 它 N N'' indicates that the pronoun ``它'' is missing before the fourth word ``吗'' in the source sentence ``你 烤 的 吗？''. More specifically, we model the probability of generating the label sequence $\bf zp$ as:
\end{CJK}
\begin{equation}\label{eqn:label}
\begin{split}
    P({\bf zp}|{\bf h}^{rec}) = \prod_{t=1}^{T}P({zp}_{t}|{\bf h}^{rec}_{t}) \\ = \prod_{t=1}^{T} g_l(zp_t, {\bf h}^{rec}_t)
\end{split}
\end{equation}
where $g_l(\cdot)$ is softmax for the ZP labeler.
As seen, we integrate the ZP generation component into the ZP translation model. There is no reliance on external ZP prediction models in decoding phase.

\paragraph{Training and Testing}

The newly introduced prediction component is trained together with the encoder-decoder-reconstructor:
\begin{equation}
\begin{split}
J(\theta, \gamma, \psi) = \argmax_{\theta, \gamma, \psi} \bigg\{ \underbrace{\log L({\bf y}|{\bf x}; \theta)}_\text{\normalsize \em likelihood} \\
+ \underbrace{\log R({\bf x} | {\bf h}^{enc}, {\bf h}^{dec}; \theta)}_\text{\normalsize \em reconstruction} \\
+ \underbrace{\log P({\bf zp} | {\bf h}^{rec}; \theta, \gamma)}_\text{\normalsize \em ZP labeling} \bigg\}
\end{split}
\end{equation}
where $\{\theta, \gamma\}$ are respectively the parameters associated with the encoder-decoder-reconstructor and the ZP prediction component.
The auxiliary prediction loss $P(\cdot)$ guides the hidden states of both the encoder-decoder and the reconstructor to embed the ZPs in the source sentence. Although the calculation of labeling loss relies on explicitly annotated labels, it is only used in training to guide the parameters to learn ZP-enhanced representations.
Benefiting from the implicit integration of ZP information, we release the reliance on external ZP prediction model in testing.

\subsection{Discourse-Aware ZP Prediction}
\label{sec:3.2}

\begin{figure}[t]
\centering
\includegraphics[width=0.48\textwidth]{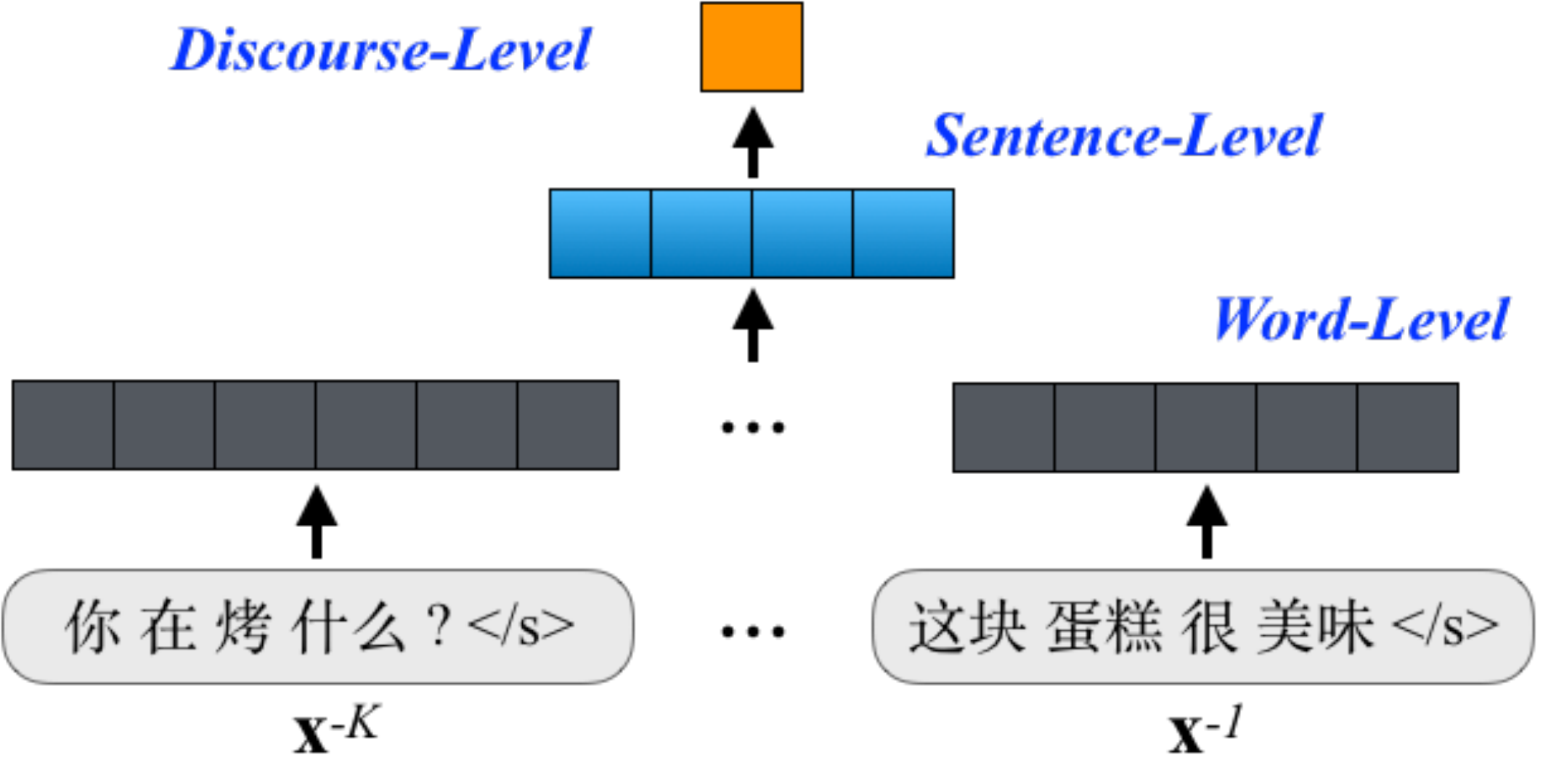}
\caption{Architecture of hierarchical neural encoder. ${\bf x}^{-K}, \dots, {\bf x}^{-1}$ are $K$ previous sentences before the current source sentence \begin{CJK}{UTF8}{gbsn}``你 烤 的 吗 ?''\end{CJK} in a text. }
\label{fig-cross-sent}
\end{figure}

Discourse information have proven useful for predicting antecedents, which may occur in previous sentences~\cite{zhao2007identification,chen2013chinese}. Therefore, we further improve ZP prediction with discourse-level context, which is learned together with the joint model.

\paragraph{Encoding Discourse-Level Context}
Hierarchical structure networks are usually used for modelling discourse context on various natural language processing tasks such query suggestion~\cite{Sordoni2015A}, dialogue modeling~\cite{Serban:2016} and MT~\cite{Wang:2017:EMNLP}. Therefore, we employ hierarchical encoder~\cite{Wang:2017:EMNLP} to encoder discourse-level context for NMT. More specifically, we use the previous $K$ source sentences ${\bf X} = \{{\bf x}^{-K}, \dots, {\bf x}^{-1}\}$ as the discourse information, which is summarized with a two-layer hierarchical encoder, as shown in Figure~\ref{fig-cross-sent}. For each sentence ${\bf x}^{-k}$, we employ a {\em word-level encoder} to summarize the representation of the whole sentence:
\begin{equation}
    {\bf h}^{-k} = \textsc{Encoder}_{word}({\bf x}^{-k})
\end{equation}
After we can obtain all sentence-level representations ${\bf H}^{X}=\{{\bf h}^{-K}, \dots, {\bf h}^{-1}\}$, we feed them into a {\em sentence-level encoder} to produce a vector that represents the discourse-level context:
\begin{equation}\label{eqn:c}
    {\bf C} = \textsc{Encoder}_{sentence}({\bf H}^{X})
\end{equation}
Here the summary $C$ consists of not only the dependencies between words, but also the relations between sentences. Following~\newcite{Voita:2018:ACL}, we share the parameters of word-level encoder $\textsc{Encoder}_{word}$ with the encoder component in standard NMT model. 
Note that, $\textsc{Encoder}_{word}$ and $\textsc{Encoder}_{sentence}$ can be implemented as arbitrary networks, such as recurrent networks~\cite{cho2014learning}, convolutional networks~\cite{Gehring:2017:ICML}, or self-attention networks~\cite{Vaswani:2017:NIPS}. In this study, we used recurrent networks to implement our $\textsc{Encoder}$.

\paragraph{Integrating Discourse into ZP Prediction}
We directly feed the discourse-level context to the reconstructor to improve ZP prediction.
Specifically, we combine the context vector and the reconstructor state:
\begin{equation}
    \widehat{\bf h}_t^{rec} = f_c ({\bf h}_t^{rec}, {\bf C})
\end{equation}
Here $f_c(\cdot)$ is a function for combining reconstructor states and the context vector, which is a simple concatenation (\textsc{Concat}) in this work.
The revised reconstructor state $\widehat{\bf h}_t^{rec}$ is then used in Equations~(\ref{eqn:rec}) and~(\ref{eqn:label}). 


\section{Experiments}

\begin{table*}[t]
\centering
\renewcommand\arraystretch{1.1}
\begin{tabular}{c||l||r|l||c|c|c}
	\multirow{2}{*}{\bf \#}    &  \multirow{2}{*}{\bf Model}        &   \multicolumn{2}{c||}{\bf Translation} &   \multicolumn{3}{c}{\bf Prediction} \\
	\cline{3-7}
	&   &   \em \#Params   &   \em BLEU    &  \em P   &  \em R   &  \em F1\\
	\hline\hline
	1   &   Baseline                &  86.7M  &   31.80   &   n/a &   n/a &   n/a\\
	\hline
	\multicolumn{7}{c}{\em External ZP Prediction~\cite{Wang:2018:AAAI}} \\
	\hline
	2   &   ~~~+ ZP-Annotated Data   &  +0M  &   32.67   &  \multirow{2}{*}{0.67}    &   \multirow{2}{*}{0.65}    &   \multirow{2}{*}{0.66}\\
	3   &   ~~~~~~~~+ Reconstruction  & +73.8M  &   35.08 &   &\\
	\hline
	\multicolumn{7}{c}{\em This Work: Joint ZP Prediction and Translation} \\
	\hline
	4   &   Joint Model    & +35.6M & 36.04$^\dag$ & 0.72   &  0.68 &  0.70\\
	5   &   ~~~+ {Discourse-Level Context} & +56.6M & \bf 37.11$^\dag$  &   0.76  & 0.77  & \bf 0.77\\
\end{tabular}
\caption{\label{tab:2} Evaluation of ZP translation and prediction on the Chinese--English data. ``\#Params'' represents the number of parameters used in different models. ``$\dag$'' indicates statistically significant difference ($p < 0.01$) from the best external ZP prediction model for translation performance. As seen, the proposed joint models improve performances in both ZP translation and prediction, over the external ZP prediction models. }
\label{tab-results}
\end{table*}

\subsection{Setup}

We conducted translation experiments on both Chinese$\Rightarrow$English and Japanese$\Rightarrow$English translation tasks, since Chinese and Japanese are pro-drop languages while English is not. For Chinese$\Rightarrow$English translation task, we used the data of auto-annotated ZPs~\cite{Wang:2018:AAAI}.\footnote{\url{https://github.com/longyuewangdcu/tvsub}.} The training, validation, and test sets contain 2.15M, 1.09K, and 1.15K sentence pairs, respectively. 
In the training data, there are 27\% of Chinese pronouns are ZPs, which poses difficulties for NMT models. 
For Japanese$\Rightarrow$English translation task, we respectively selected 1.03M, 1.02K, and 1.02K sentence pairs from Opensubtitle2016\footnote{\url{ http://www.opensubtitles.org}.} as training, validation, and test sets~\cite{tiedemann2012parallel}. 
We used case-insensitive 4-gram NIST BLEU \citep{Papineni:2002} as evaluation metrics, and {\em sign-test} \citep{Collins05} to test for statistical significance.

To make fair comparison with~\newcite{Wang:2018:AAAI}, we also implemented our approach on top of the RNN-based NMT model, which incorporates dropout \cite{hinton2012improving} on the output layer and improves the attention model by feeding the most recently generated word. For training the models, we limited the source and target vocabularies to the most frequent 30K words for Chinese$\Rightarrow$English and 20K for Japanese$\Rightarrow$English. 
Each model was trained on sentences of length up to a maximum of 20 words with early stopping. Mini-batches were shuffled during processing with a mini-batch size of 80. The dimension of word embedding was 620 and the hidden layer size was 1,000. We trained for 20 epochs using Adadelta~\cite{zeiler2012adadelta}, and selected the model that yielded best performances on validation sets. For training the proposed models, the hidden layer sizes of hierarchical model and reconstruction model are 1,000 and 2,000, respectively. We modeled previous three sentences as discourse-level context.\footnote{We followed \newcite{Wang:2017:EMNLP} and \newcite{Tu:2018:TACL} to use 3 previous sentences as discourse context.}



\subsection{Results on Chinese$\Rightarrow$English Task}
\label{sec:4.2}


Table~\ref{tab-results} lists the performance of ZP translation and prediction on Chinese$\Rightarrow$English data.

The baseline (Row 1) is trained on the standard NMT model using the original parallel data ({$\bf x$}, {$\bf y$}). In addition, we implemented two comparative models (Row 2-3), which differ with respect to the training data used. The ``+ ZP-Annotated Data'' model was still trained on standard NMT model but using new training instances ({$\bf \hat{x}$}, {$\bf y$}) whose source-side sentences are auto-annotated with ZPs. The ``+ Reconstruction'' is the best model reported in~\newcite{Wang:2018:AAAI}, which employs two reconstructors to reconstruct the $\bf \hat{x}$ from hidden representations of encoder and decoder. At decoding time, ZPs can not be annotated by alignment method since target sentences are not available. Thus, source sentences are annotated by an external ZP prediction model, which is trained on monolingual training instances $\bf \hat{x}$. Finally, we evaluated two proposed models (Row 4-5) which are introduced in Section~\ref{sec:3.1} and \ref{sec:3.2}, respectively.

\paragraph{Translation Quality}
Benefiting from the explicitly annotated ZPs in the source language, the ``+ ZP-Annotated Data'' model (Row 2) outperforms the baseline system built on the original data where the pronouns are missing (\ie +0.87 BLEU point). This illustrates that explicitly recalling translation of ZPs at training time helps produce better translations.
Furthermore, the ``+ Reconstuction'' approach (Row 3) respectively outperforms the baseline and ``+ ZP-Annotated Data'' models by +3.28 and +2.41 BLEU points, which indicates that explicitly handling ZPs with reconstruction model can better address ZP problems. 

The proposed models consistently outperform other models in all cases, demonstrating the superiority of the joint learning of ZP prediction and translation.
Specifically, the ``Joint Model'' (Row 4) significantly improves translation performance by +4.24 over baseline model. In addition, this joint approach also outperforms two comparative models ``+ ZP-Annotated Data'' and ``+ Reconstruction'' by +3.37 and +0.96 BLEU points, respectively. We attribute the improvement over external ZP prediction to: 1) releasing the reliance on external ZP prediction models can greatly alleviate error propagation problems; and 2) joint learning of ZP prediction and translation is able to guide the related parameters to learn better latent representations. Furthermore, introducing discourse-level context (Row 5) accumulatively improves translation performance, and significantly outperform the joint model by +1.07 BLEU points. 



More parameters may capture more information, at the cost of posing difficulties to training. ~\newcite{Wang:2018:AAAI} leverage two separate reconstructors with hidden state size being 2000 and 1000 respectively. 
Accordingly, their models introduce a large number of parameters. In contrast, we set the hidden size of the reconstructor be 1000, which greatly reduce the newly introduced parameters (+35.6M vs. +73.8M). Modeling discourse-level context further introduces +21M new parameters, which is reasonable comparing with previous work. 
Our best model variation outperform that of external ZP prediction by over 2 BLEU points with less parameters (143.3M vs. 160.5M), showing that the improvements are attributed to the stronger modeling capacity rather than more parameters.



\begin{table}[t]
\centering
\renewcommand\arraystretch{1.1}
\begin{tabular}{l|c|c}
	\bf Model & \bf BLEU & \bf $\bigtriangleup$ \\
	\hline
	Baseline    &   19.94   &   --\\
	External ZP Prediction & 20.86   & +0.92  \\
	\hline
	Joint Model & 21.39 & +1.45 \\
	~~+ Discourse-Level Context & {\textbf{22.00}}  &   +2.06 \\
\end{tabular}
\caption{Translation quality on Japanese--English data. As seen, the proposed models can also significantly improve translation performance, which shares the same trend with that on Chinese--English translation.}
\label{tab-results-jaen}
\end{table}

\paragraph{ZP Prediction Accuracy}
The joint model improves prediction accuracy as expected, which we attribute to the leverage of useful translation information. Incorporating the discourse-level context further improves ZP prediction, and the best performance is 11\% higher than external ZP prediction model. These results confirm our claim that joint learning of ZP prediction and translation can benefit both components by allowing them to interact with each other.




\subsection{Results on Japanese$\Rightarrow$English Task}

Table \ref{tab-results-jaen} lists the results. We compare our models and the best external ZP prediction approach. As seen, our models also significantly improve translation performance, demonstrating the effectiveness and universality of the proposed approach.

This improvement on Japanese$\Rightarrow$English translation is lower than that on Chinese$\Rightarrow$English, showing that ZP prediction and translation are more challenging for Japanese. The reason may be two folds: 1) Japanese language has a larger number of pronoun variations borrowed from archaism, which leads to more difficulties in learning ZPs; 2) Japanese language is subject-object-verb (SOV) while English has subject-verb-object (SVO) structure, and this poses difficulties for ZP annotation via alignment method.



\subsection{Analysis}
\label{sec:4.3}

We conducted extensive analyses on Chinese $\Rightarrow$English to better understand our models in terms of the effect of external ZP annotation and different types of ZPs errors.

\begin{table}[t]
\centering
\renewcommand\arraystretch{1.1}
\begin{tabular}{l|c|c|c}
	\multirow{2}{*}{\bf Model} & \multicolumn{3}{c}{\bf ZP-Annotated Input} \\
	\cline{2-4}
	    &   \checkmark  &   \texttimes  &   $\bigtriangledown$ \\
	\hline \hline
 	Baseline          &   \multicolumn{2}{c|}{31.80}   &   -- \\
 	\hdashline
	External ZP Predict. &   35.08 &   34.02   &   -1.06\\
	\hline
	Joint Model      &    36.04   &  35.93  &  -0.11\\
	~~~+ Discourse   &    \bf 37.11   &  \bf 36.51  &  -0.60\\
\end{tabular}
\caption{\label{tab:rec} Translation results when no ZP-annotated input is used in decoding by {\em removing the reconstructor component}. ``$\bigtriangledown$'' denotes the performance gap between whether using the annotated input (``\checkmark'') or not (``\texttimes'').}
\end{table}

\paragraph{Reliance on Externally ZP-Annotated Input}
Some researchers may argue that previous approaches~\cite{Wang:2018:AAAI} are also able to release the reliance of externally annotated input by removing the reconstructor component. Table~\ref{tab:rec} lists the results. Without ZP-annotated input in decoding, all approaches can still outperform the baseline model, by benefiting better intermediate representations that contain necessary ZP information. Compared with reconstruction-based models, however, removing the reconstruction components leads to decrease on translation quality. As seen, the BLEU score of best ``External ZP prediction'' model dramatically drops by -1.06 points, showing that this approach is heavily dependent on the results of external ZP annotations. The performances of proposed models only decrease by -0.1$\sim$-0.6 BLEU point. It indicates that our models are compatible with the standard encoder-decoder-reconstructor framework, thus enjoy an additional benefit of re-scoring translation hypotheses in testing with reconstruction scores. All the results together prove the superiority of the proposed unified framework for ZP translation.

\begin{table}[t]
\centering
\renewcommand\arraystretch{1.1}
\begin{tabular}{l|c|c}
	\bf Model & \bf BLEU & \bf $\bigtriangleup$ \\
	\hline
	Baseline                        &  31.80   &   --\\
    ~~~+ Discourse$\Rightarrow$Decoder   &  32.34   &   +0.54\\
    \hline
	Baseline + ZP-Anno. & 32.67   &   \\
	~~~+ Discourse$\Rightarrow$Decoder & 32.55 &  -0.12\\
	\hline
	Joint Model   &   36.04  &   --\\
	~~~+ Discourse$\Rightarrow$Decoder   &  34.66   &   -1.38\\
\end{tabular}
\caption{Translation results when transforming the contextual representation to decoder of different models. Incorporating discourse-level context does not always lead to improvement of translation performance.} 
\label{tab-results-discourse}
\end{table}

\paragraph{Effect of Discourse-Level Context}
Recent studies revealed that inter-sentential context can implicitly help to tackle anaphora resolution in NMT architecture~\cite{jean2017neural,bawden2018evaluating,Voita:2018:ACL}. Some may argue that document-level architectures are strong enough to alleviate ZP problems for NMT. To answer this concern, we compared with ``+ Discourse$\Rightarrow$Decoder'' models, which transform the contextual representation to the decoder part of different models. In this way, the discourse-level context can benefit both the generation of translation and ZP prediction.

As shown in Table~\ref{tab-results-discourse}, directly incorporating inter-sentential context into standard NMT model (one of document-level NMT architectures) can improve translation quality by +0.54 BLEU point than baseline. However, this integration mechanism does not work well in ``Baseline + ZP-Annotation'' and our ``Joint'' models, which decreasing by -0.12 and -1.38 BLEU points, respectively. One potential problem with this strategy is that the propagation path is longer: ${\bf C} \rightarrow {\bf h}^{dec} \rightarrow {\bf h}^{rec} \rightarrow {\bf zp}$, which may suffer from the vanishing effect. 
This also confirms our hypothesis that discourse-level context benefits ZP prediction more than ZP translation. Therefore, we incorporate the discourse-level context into reconstructor instead of the decoder.

\begin{table}[t]
\renewcommand\arraystretch{1.1}
\centering
\begin{tabular}{l|c|ccc|c}
{\bf Model} & \bf Error & \bf Sub. & \bf Obj. & \bf Dum. & \bf All\\ 
\hline
\textsc{Base.} & Total  & 112 & 41 & 45 & 198 \\ \hline
\multirow{3}{*}{\textsc{Exte.}} 
& Fixed & 50 & 34 & 33  & 117 \\
& New & 11 & 14 & 7 & 32 \\ 
& Total &  73   &   21  &   19  &   113\\
\hline
\multirow{3}{*}{\textsc{Join.}} 
& Fixed & 61 & 35 & 37  & 133 \\
& New & 8 & 11 & 7 & 26 \\
& Total &  59 & 17  &  15  &  91\\
\hline
\multirow{3}{*}{\textsc{~~+Dis.}} 
& Fixed & 70 & 39 & 38  & 147 \\
& New & 7 & 9 & 7 & 23 \\
& Total & \bf 49  &  \bf 11  &  \bf 14  &  \bf 74 \\ 
\end{tabular}
\caption{\label{tab:mannual} Translation error statistics. The ZP types ``Sub.'', ``Obj.'' and ``Dum.'' denote errors caused by subjective, objective and dummy pronouns, respectively. The models ``Base.'', ``Exte.'', ``Join.'' and ``+Dis.'' denote ``Baseline'', ``+ Reconstruction'', ``Joint Model'' and ``+ Discourse-Level context'' models. {\bf Bold} numbers denote the least errors in each category.} 
\end{table}

\paragraph{Manual Evaluation on Translation Errors}
We finally investigate how the proposed approaches improve the translation by human evaluation. We randomly select 500 sentences from the test set. As shown in Table~\ref{tab:mannual}, we count how many translation errors caused by different types of ZPs (\ie ``Subjective'', ``Objective'' and ``Dummy''\footnote{In pro-drop languages, it is used to fulfill the syntactical requirements without providing explicit meaning (e.g. ``it'').}) are fixed (``Fixed'') and newly generated (``New'') by different models.

All the models can fix different amount of ZP problems in terms of completeness and correctness, which is consistent with the translation results reported in Table~\ref{tab-results}. This confirms that our improvement in terms of BLEU scores indeed comes from alleviating translation errors caused by ZPs. Among them, the proposed model ``\textsc{+Dis.}'' performs best, which fixes 74\% of the ZP errors, and only introduces 12\% of new errors. 

In addition, we found that subjective ZPs are more difficult to predict and translate since they usually occur in imperative sentences, and ZP prediction needs to understand intention of speakers. The ``\textsc{Exte.}'' model only fixes 45\% of subjective ZP errors but made 10\% new errors by predicting wrong ZPs. However, the proposed joint model works better, which fixes 54\% error with only introducing 7\% new errors. Predicting objective ZPs needs inter-sentential context, thus our ``\textsc{+Dis.}'' model is able to fix more objective ZP errors (95\% vs. 82\%) by introducing less new errors (22\% vs. 34\%) than ``\textsc{Exte.}''.

\begin{CJK}{UTF8}{gbsn}
\begin{table}[t]
\renewcommand\arraystretch{1.1}
\centering
\begin{tabular}{r|l}
\hline
\multicolumn{2}{c}{\bf Fixed Error} \\
\hline
\textsc{Pre.} & 等 我 搬进 来, 能 买 台 电视 吗?\\
\textsc{Inp.} & 当然 可以, 乔伊 不让 {(你)} 买 {(它)}?\\ 
\textsc{Ref.} & Sure. Joey wouldn't let you buy it?\\ 
\textsc{Exte.} & Of course. Sure, Joey won't get {\em \color{blue}it}?\\ 
\textsc{Join.} & Sure. Joey won't let {\em \color{blue}us} buy {\em \color{blue}one}?\\
\textsc{+Dis.} & Sure. Joey wouldn't let {\bf \color{red}you} buy {\bf \color{red}it}?\\
\hline
\multicolumn{2}{c}{\bf Non-Fixed Error} \\
\hline
\textsc{Pre.} & 我 和 露西 只是 要 搬 到 对门。\\
\textsc{Inp.} & 我们 一 分手 {(我)} 就 搬 回去。\\
\textsc{Ref.} & Once we broke up, I'll move back.\\
\textsc{Exte.} & Once we broke up, {\em \color{blue}she}'ll move back.\\
\textsc{Join.} & Once we broke up, {\em \color{blue}we} moved back.\\
\textsc{+Dis.} & Once we broke up, {\em \color{blue}we}'ll move back.\\
\hline
\end{tabular}
\caption{\label{fig-example} Example translations where pronouns in brackets are dropped in original inputs (``\textsc{Inp.}'') but labeled by humans according to references (``\textsc{Ref.}'') and previous sentence (``\textsc{Pre.}''). We italicize some {\em \color{blue} mis-translated} errors and highlight the {\bf \color{red} correct} ones in bold. }
\end{table}
\end{CJK}

\paragraph{Case Study}
\begin{CJK}{UTF8}{gbsn}
Table~\ref{fig-example} shows two typical examples, of which pronouns are mistakenly translated by the strong baseline (``External ZP Prediction'') model~\cite{Wang:2018:AAAI} while fixed by our model and failed to be fix. In ``Fixed Error'' case, the dropped word ``它 (\textit{it})'' is an anaphoric ZP whose antecedent is the noun ``电视 (\textit{television})'' in previous sentence while the dropped word ``你 (\textit{you})'' is a non-anaphoric ZP that depends upon speaker or listener. As seen, our ``\textsc{Join.}'' model performs better than the ``\textsc{Exte.}'' model because two ZP positions are syntactically recalled in the target side, showing that the joint approach have better capability of utilizing intra-sentential information for identifying ZPs. Besides, our ``\textsc{+Dis.}'' model can semantically fix the error by predicting correct ZP words, demonstrating that inter-sentential context can aid to recovering such complex ZPs. However, as shown in ``Non-Fixed Error'' case, there are still some ZPs can not be precisely predicted due to the misunderstanding of intentions of utterances. Thus, exploiting dialogue focus for ZP translation is our future work~\cite{rao2015dialogue}.

\end{CJK}


\section{Related Work}

\paragraph{ZP Prediction and Translation}

ZP resolution is a challenging task which needs lexical, syntactic, discourse knowledge. Previous studies have been conducted to improves the performance of ZP resolution for different pro-drop languages~\cite{kong2010tree,chen2013chinese,park2015zero,yin2017chinese}. However, directly using results of external ZP resolution systems for translation task shows limited improvements~\cite{chung2010effects,Nagard:2010:ACL,Taira:2012:SSSST,xiang2013enlisting}, since such external systems are trained on small-scale data that is non-homologous to MT. 
To overcome the data-level gap, \newcite{Wang:2016:NAACL} proposed an automatic approach of ZP annotation by utilizing an alignment matrix from a large parallel data. By using the translation-oriented ZP corpus, they exploited different approaches to alleviate ZP problems for translation models~\cite{Wang:2016:NAACL,Wang:2018:AAAI,Wang:2018:EMNLP}. Note that \newcite{Wang:2018:EMNLP} also explored to address the problem of error propagation by jointly predicting ZP words given ZP position information. However, this method still relies an external model that predicting ZP positions at decoding time. 
Instead, this work proposes a unified model without any additional ZP annotations in decoding, thus release reliance on external ZP prediction in practice.


\paragraph{Discourse-Aware NMT}
Recent years, context-aware architecture has been well studied for NMT~\cite{Wang:2017:EMNLP,jean2017does,Tu:2018:TACL}.
\newcite{Wang:2017:EMNLP} proposed hierarchical recurrent neural networks to summarize inter-sentential context from previous sentences and then integrate it into a standard NMT model with difference strategies. \newcite{jean2017does} introduced an additional set of an encoder and attention to encode and select part of the previous source sentence for generating each target word. Besides, \newcite{Tu:2018:TACL} proposed to augment NMT models with a cache-like memory network, which stores the translation history in terms of bilingual hidden representations at decoding steps of previous sentences. They also evaluated the above three models on different domains of data, showing that the hierarchical encoder performs comparable with the multi-attention model. More recently, some researchers began to investigate the effects of context-aware NMT on cross-lingual pronoun prediction~\cite{jean2017neural,bawden2018evaluating,Voita:2018:ACL}. They mainly exploited general anaphora in non-pro-drop languages such as English$\Rightarrow$Russian.





\section{Conclusion}
\label{sec:6}


In this work, we proposed a unified model to learn jointly predict and translate ZPs by leveraging multi-task learning. We also employed hierarchical neural networks to exploit discourse-level information for better ZP prediction. 
Experimental results on both Chinese$\Rightarrow$English and Japanese$\Rightarrow$English data show that the two proposed approaches accumulatively improve both the translation performance and ZP prediction accuracy. Our models also outperform the existing ZP translation models in previous work, and achieve a new state-of-the-art on the widely-used subtitle corpus. Manual evaluation confirms that the performance improvement comes from the alleviation of translation errors, which are mainly caused by subjective, objective as well as discourse-aware ZPs. 

There are two potential extensions to our work. First, we will evaluate our method on other implication phenomena (or called unaligned words~\cite{takeno2017controlling}) such as tenses and article words for NMT. Second, we will investigate the impact of different context-aware models on ZP translation, including multi-attention~\cite{jean2017neural} and context-aware Transformer\cite{Voita:2018:ACL}.

\bibliography{emnlp-ijcnlp-2019}
\bibliographystyle{acl_natbib}

\end{document}